\documentclass{article}
\usepackage[affil-it]{authblk}
\usepackage[bottom]{footmisc}

\usepackage{geometry}
\makeatletter
\def\@maketitle{%
  \newpage
  \null
  \vskip 2em%
  \begin{center}%
  \let \footnote \thanks
    {\Large\bfseries \@title \par}%
    \vskip 1.5em%
    {\normalsize
      \lineskip .5em%
      \begin{tabular}[t]{c}%
        \@author
      \end{tabular}\par}%
    \vskip 1.5em%
    {\normalsize \@date}%
  \end{center}%
  \par
  \vskip 1.5em}
\makeatother

\usepackage{doc}
\usepackage{subcaption}

\usepackage{url}

\usepackage{xcolor}

\usepackage{graphicx}
\usepackage{epstopdf}
\title{Comment on \textit{Biologically inspired protection of deep networks from adversarial attacks}}

\author[1,3]{Wieland Brendel}

\author[1,2,3,4]{Matthias Bethge}%

\affil[1]{Werner Reichardt Center for Integrative Neuroscience, University of T\"ubingen, Germany}
\affil[2]{Max Planck Institute for Biological Cybernetics, T\"ubingen, Germany}
\affil[3]{Bernstein Center for Computational Neuroscience, T\"ubingen, Germany}
\affil[4]{Institute for Theoretical Physics, University of T\"ubingen, Germany}

\date{Dated: \today}

\begin{document}

\maketitle

\textbf{A recent paper \cite{1703.09202} suggests that Deep Neural Networks can be protected from gradient-based adversarial perturbations by driving the network activations into a highly saturated regime. Here we analyse such saturated networks and show that the attacks fail due to numerical limitations in the gradient computations. A simple stabilisation of the gradient estimates enables successful and efficient attacks. Thus, it has yet to be shown that the robustness observed in \cite{1703.09202} is not simply due to numerical limitations.}

Evaluating the robustness of neural networks is difficult. One core reason is the ambiguity between network robustness and deficiencies of the adversarial attack: a network might just appear robust because the core assumptions of the chosen attacker are not met. This is particularly obvious for gradient-based adversarial attacks which inadvertently rely on stable gradient estimates. Even in cases without gradients, however, many other adversarial attacks may still succeed, including attacks that do not use gradient information like \cite{clune,papernot} or gradient-based attacks using estimates of the gradient.

A recent paper \cite{1703.09202} suggests that highly saturated  deep neural networks (DNNs) might be robust against gradient-based adversarial perturbations. We here show that the observed robustness is likely a side-effect of numerical limitations that arise in the high-saturation limit and which prevent stable computations of the gradient. These limitations can be lifted by a simple and stable estimate of the gradients.

In a first step we tried to reproduce the results of \cite{1703.09202} by training a three-layer Multi-layer Perceptron (MLP) together with the proposed saturation penalty. This penalty pushes the hidden-layer activations of the network into the saturated parts of the non-linearity (zero and one for sigmoid, zero for ReLU). We report the classification accuracy for both the vanilla and the saturated network for normal and adversarial images in Table \ref{table:accuracy}. In agreement with \cite{1703.09202} the performance of the MLP with and without penalty is almost identical, but the robustness to a simple adversarial attack with the fast-gradient sign method (FGSM) drastically increases for the latter. We also verified that the weight and activation distribution qualitatively match the results in \cite{1703.09202}, Figure \ref{fig:parameterDistribution}.

\begin{table}[b]
\begin{tabular}{ |p{1,6cm}|p{1,3cm}|p{2cm}|p{2cm}|p{1,3cm}|p{2cm}|p{2cm}|}
 \hline
 \multicolumn{1}{|c|}{Training} & \multicolumn{3}{|c|}{Sigmoid MLP} & \multicolumn{3}{|c|}{ReLU MLP} \\
 \hline
 & Plain & naive FGSM & stable FGSM & Plain & naive FGSM & stable FGSM \\
 \hline
 Vanilla   & 98\% &  2,5\% & - & 98,7\% & 0,2\% & -\\
 Saturated & 97,2\% & 96,6\% & 1,7\% & 98,1\% & 98,0\% & 8,4\% \\
 \hline
\end{tabular}
\caption{\label{table:accuracy}Adversarial robustness of vanilla and saturated sigmoidal and ReLU MLPs. While the naive application of FGSM fails to generate suitable adversarial examples, a slightly modified FGSM based on a more stable gradient estimate is still highly successful.}
\end{table}

\begin{figure}[t]
  \centering
    \includegraphics[width=0.8\textwidth]{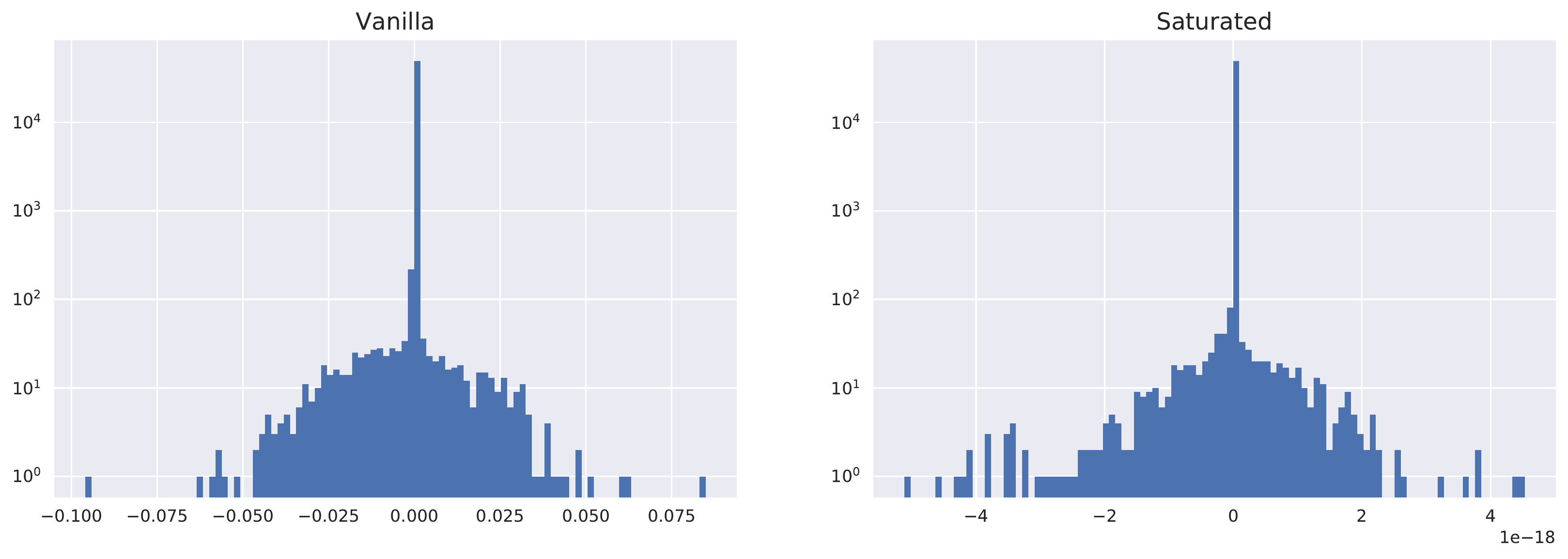}
  \caption{\label{fig:gradients}Histogram over the elements of the gradients of the input image with respect to the cross-entropy loss (the direction of the adversarial perturbation) for both the vanilla sigmoid MLP (left) and the saturated sigmoid MLP (right). In the saturated network more then 98\% of the gradient elements are exactly zero while the rest is sixteen orders of magnitude smaller then in the vanilla network.}
\end{figure}

In highly saturated networks the gradients of the loss with respect to the input are either exactly zero or numerically unstable. Using these unreliable values from the saturated network directly without numerical stabilisation is likely to fail. In Figure \ref{fig:gradients} we plot the distribution of elements of the gradients. In the saturated network more then 98,2\% of the gradient elements are exactly zero, compared to none in the vanilla network. At the same time, all non-zero elements are sixteen orders of magnitude smaller the gradient elements of the vanilla network, suggesting that the residual gradients of the saturated network are due to rounding errors or are susceptible to numerical instabilities. For 97,9\% of the images exactly all elements of the gradient are zero. In this case FGSM applies absolutely no perturbation to the corresponding image and the attack is inadvertently unsuccessful. For all other images the FGSM attack was successful in 62\% of the cases.

\begin{figure}[b]
  \centering
    \includegraphics[width=0.8\textwidth]{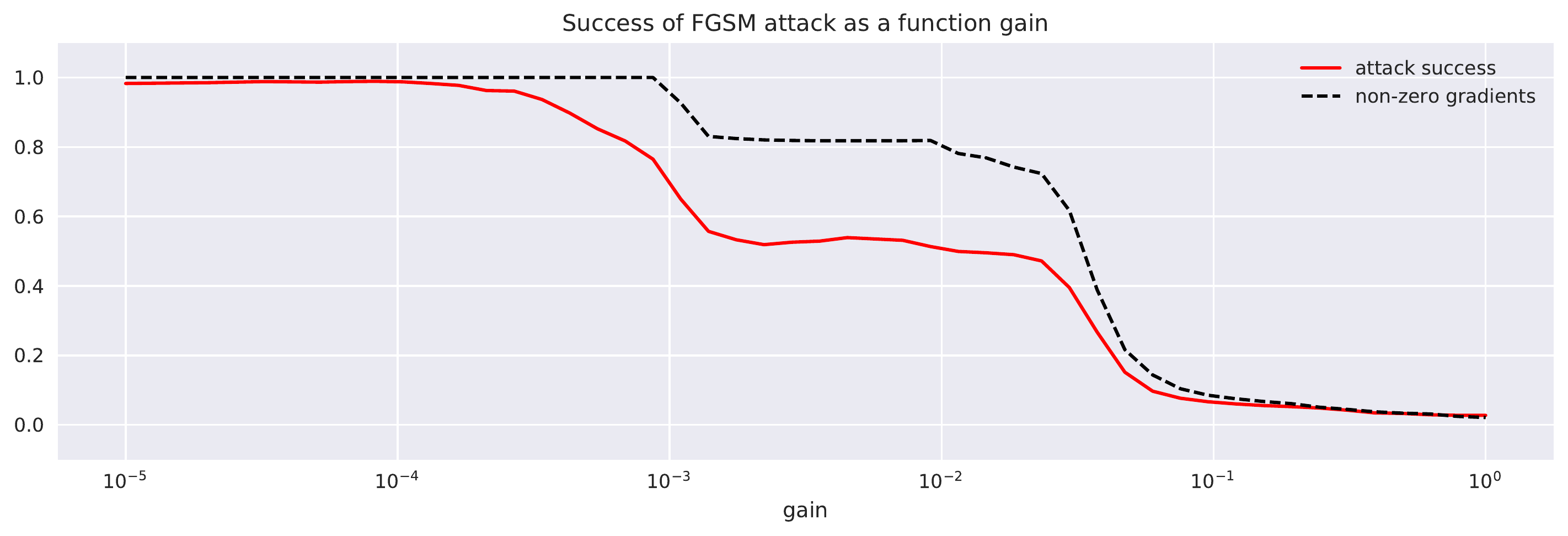}
  \caption{\label{fig:gradient_attack} The success of the FGSM attack clearly reflects the ratio of non-zero gradients. Networks with different gain are only used to generate adversarial images using the FGSM method. The accuracy by which the adversarials fool the saturated network (gain = 1) is plotted in red. The success of FGSM is highly correlated with the ratio of non-zero gradients (black).}
\end{figure}

The overall success rate of FGSM is directly related to the number of zero-valued gradients. In Figure \ref{fig:gradient_attack} we plot the success of FGSM and the ratio of non-zero gradients as a function of the gain. For gains below $10^{-3}$ FGSM is highly successful (as evaluated on the saturated sigmoid MLP with gain 1). For larger gains, however, zero-valued gradients start to appear. Unsurprisingly, the success rate of FGSM strongly decreases with the number of zero-valued gradients.

It is straight-forward to attack saturated networks through a simple trick that allows more stable gradient estimates. To this end we note that in highly saturated networks a modest reduction in the gain of the sigmoids barely change the activations (they will still be close to zero and one) but can significantly increase the numerical stability of the gradient estimates. We then use this gradient to generate an adversarial example according to the usual FGSM procedure and use it as an input to the original saturated network. As can be seen in Table \ref{table:accuracy}, this simple modification of FGSM is highly successful in fooling the saturated network. For saturated ReLU MLPs we observed a saturation of the softmax and devised a similar attack by down-scaling the activations of the readout layer (Table \ref{table:accuracy}).

Taken together, we demonstrated that the robustness observed in \cite{1703.09202} likely originates from the numerical instabilities of gradient computations in highly saturated networks. A simple stabilisation of the gradient computations still allow standard gradient-based adversarial attack methods to succeed. While we cannot say with certainty that the same attack succeeds for the networks analysed in \cite{1703.09202} (without access to the source code we cannot exclude unintended differences in the implementation and the network parameters), it has yet to be shown that the robustness observed in \cite{1703.09202} does not originate from numerical limitations. More generally, our findings highlight the critical importance of choosing the most suitable methods to challenge the robustness of the network.

\bibliography{main}
\bibliographystyle{unsrt}

\appendix
\renewcommand\thefigure{S\arabic{figure}}
\setcounter{figure}{0}

\begin{figure}[b]
\centering
   \begin{subfigure}[b]{1\textwidth}
   \includegraphics[width=1\linewidth]{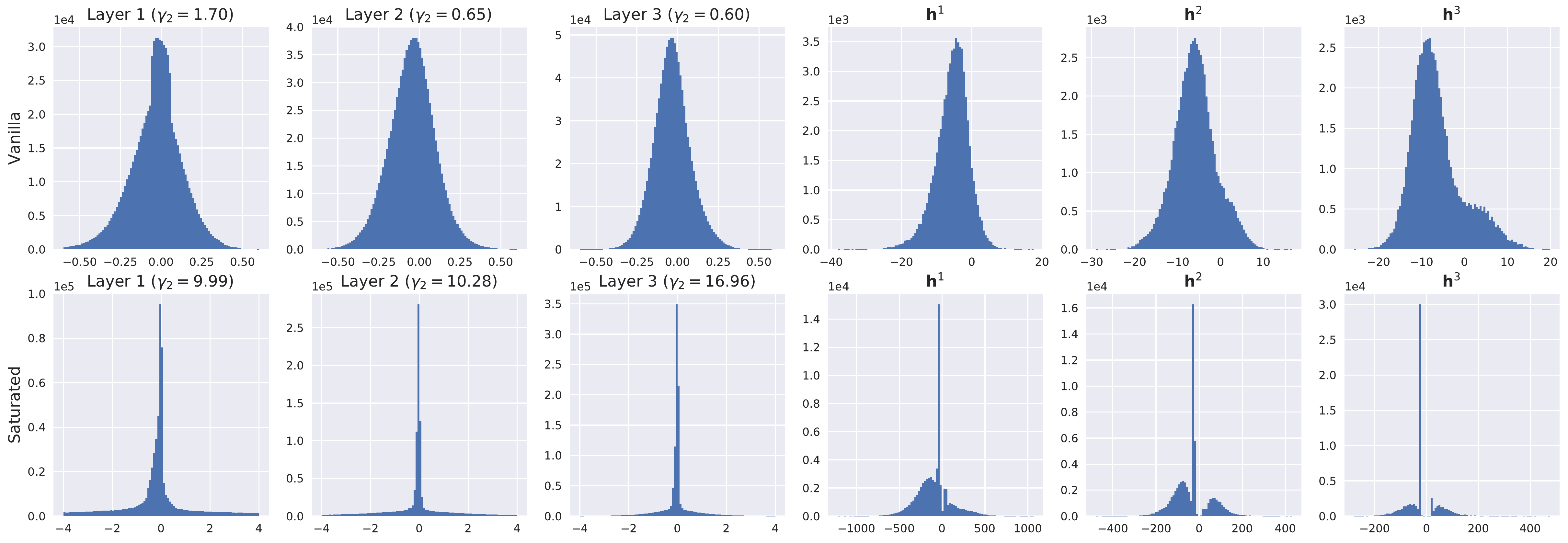}
   \caption{}
   \label{fig:Ng1} 
\end{subfigure}

\begin{subfigure}[b]{1\textwidth}
   \includegraphics[width=1\linewidth]{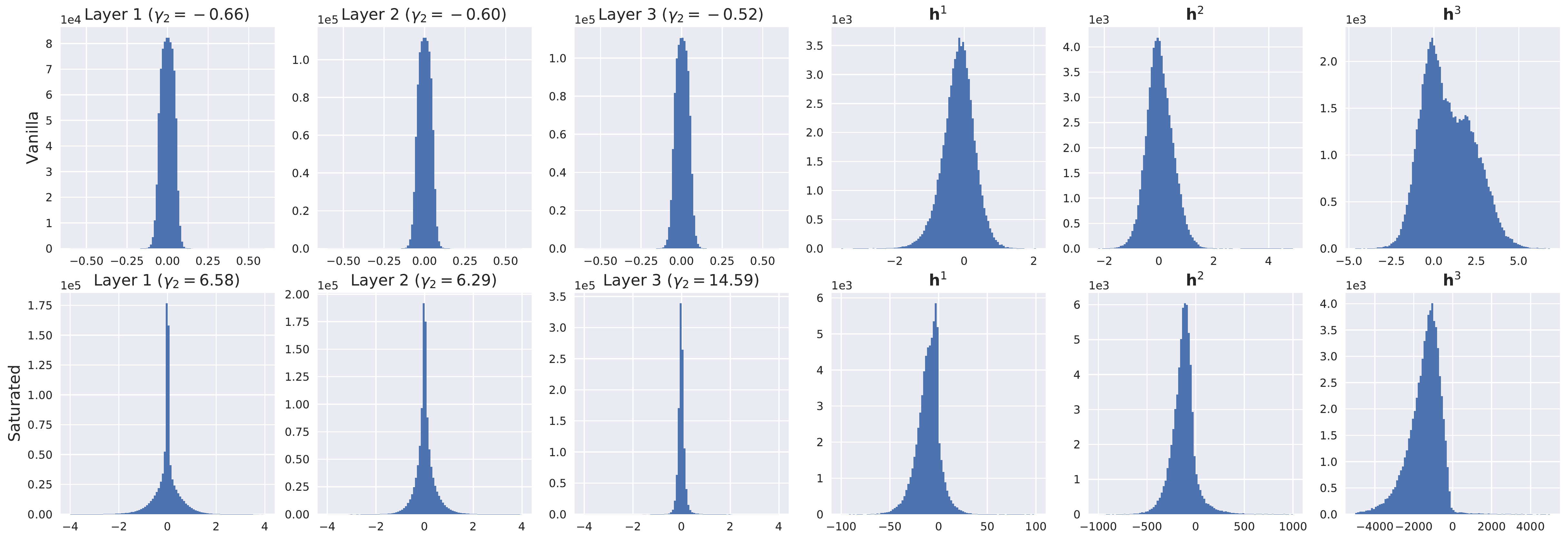}
   \caption{}
   \label{fig:Ng2}
\end{subfigure}

\caption{\label{fig:parameterDistribution} (a) \textbf{Sigmoid MLP} Weight and activation distribution for both the vanilla (top) and saturated (bottom) network. We observe a qualitatively similar increase in the kurtosis of the weights and the bimodality of the activations as in \cite{1703.09202}. (b) \textbf{ReLU MLP} Same as (a) but for ReLU nonlinearities. Similar to \cite{1703.09202} the activations are not bimodal as in the sigmoid MLP but feature a high kurtosis.}
\end{figure}

\end{document}